\pdfoutput=1

\documentclass[11pt]{article}

\usepackage[]{acl}

\usepackage{times}
\usepackage{latexsym}

\usepackage[T1]{fontenc}

\usepackage[utf8]{inputenc}

\usepackage{diagbox}
\usepackage{xurl}

\usepackage{makecell}

\usepackage{microtype}
\usepackage{amsthm}
\usepackage{verbatim}
\theoremstyle{definition}
\newtheorem*{definition*}{Definition}

\usepackage[normalem]{ulem}
\usepackage{multirow}

\usepackage[normalem]{ulem}

\usepackage{graphicx}

\title{Human-Centric Research for NLP: \\Towards a Definition and Guiding Questions}

  \author{$\forall$\textsuperscript{*}, Bhushan Kotnis, Kiril Gashteovski, Julia Gastinger, Giuseppe Serra,\\ \textbf{Francesco Alesiani, Timo Sztyler, Ammar Shaker, Na Gong, Carolin Lawrence, Zhao Xu}\\
	NEC Laboratories Europe\\	
	Heidelberg, Germany \\
	\texttt{\{human-ai-centric\}@neclab.eu} \\}
  
\begin{document}
\maketitle
\begingroup\renewcommand\thefootnote{*}
\footnotetext{All authors contributed equally, order has been randomized (see \url{https://bit.ly/3eS6UUR}).}
\endgroup
\begin{abstract}
With Human-Centric Research (HCR) we can steer research activities so that the research outcome is beneficial for human stakeholders, such as end users. But what exactly makes research \textit{human-centric}? We address this question by providing a working definition and define how a research pipeline can be split into different stages in which human-centric components can be added. Additionally, we discuss existing NLP with HCR components and define a series of \textit{guiding questions}, which can serve as starting points for researchers interested in exploring human-centric research approaches. We hope that this work would inspire researchers to refine the proposed definition and to pose other questions that might be meaningful for achieving HCR.
\end{abstract}

\section{Introduction}
Research advances are driven by curiosity and the need to improve human life. Often research addresses fundamental core challenges. Other times research is more application oriented. With the recent significant progress in NLP, the NLP community has the opportunity to address more and more application oriented research questions. However, such research is not necessarily human-centric: the research questions and the methods for answering them are often entirely designed and implemented by the researchers and their intuition. 

In contrast, Human-Centric Research (HCR) aims to ensure that \textit{human stakeholders} benefit from the outcome of the research. The stakeholders are people that have a stake in the research outcome and may include researchers,
end users, data collectors, feedback providers or domain experts. We use the term \textit{external stakeholders} to refer to stakeholders other than the researchers. In HCR, the goal is to place \emph{all} stakeholders at the center of research, not only the researchers themselves. 
Not involving external stakeholders introduces the risk of spending great amounts of time and resources in solving problems that might turn out to be irrelevant in practical settings. With HCR, we would like to ensure that external stakeholders benefit from the outcome of the research by directly involving them in the research process.

\begin{figure}[t]
    \centering
    \includegraphics[scale=0.45,trim={5.5cm 5.3cm 5.5cm 5.3cm},clip]{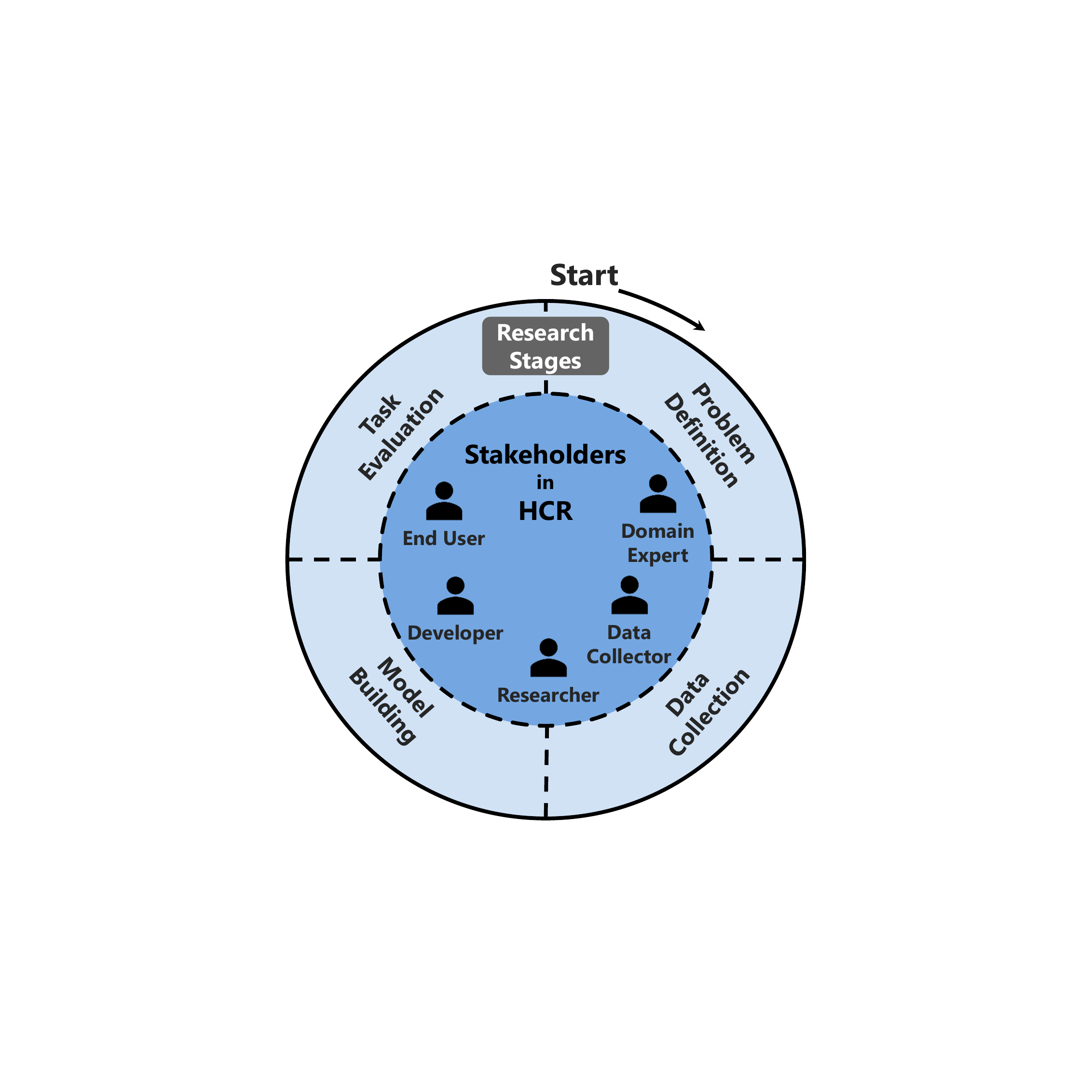}
    \caption{Human-Centric Research (HCR) should involve stakeholders other than the researchers themselves. We define four stages in the research pipeline where other stakeholders can be involved.}
    \label{fig:dt-nlp}
    \vspace{-0.5em}
\end{figure}

Human-centric research components are used increasingly in the NLP community. For example in evaluation \cite{ribeiro-etal-2020-beyond,hong2021planning}, design \cite{park2021facilitating, wang-etal-2021-putting, iskender-etal-2021-towards} and exploration \cite{lertvittayakumjorn2021explanation,heuer-buschek-2021-methods} of NLP models in various tasks such as machine translation \cite{nekoto-etal-2020-participatory}, dialogue systems \cite{Xiao2020chatbot}, text summarization \cite{smith2018closing,hsu-tan-2021-decision,passali-etal-2021-towards}, natural language generation \cite{tintarev2016personal,clark2018creative,akoury-etal-2020-storium}, information extraction \cite{gashteovski-etal-2020-aligning} and other applications \cite{rello2015spellchecker}. 

As a community, however, we lack a clear definition of what HCR means. With a clear definition it would be easier for researchers to find more human-centric research questions and to better communicate in what way their research is human-centric. To tackle these open problems, we make the following contributions. First, we formulate a working definition for the term ''human-centric research`` and define different stages of research which can be made more human-centric (Section \ref{sec:definition}). For an overview of the research stages see Figure \ref{fig:dt-nlp}. Second, we review recent literature in the field of NLP with respect to HCR and formulate and discuss a series of guiding research questions for future HCR projects. Finally, we present one example case study that showcases how the guiding questions can be adapted for particular NLP tasks (Sec.~\ref{sec:case-study-summarization}).

\section{Definition \& Research Stages}\label{sec:definition}
Human-centric AI (HCAI) has attracted increasing attention \cite{shneiderman2022human}, but in the broader research community it is still an open question how to best conduct HCR. Based on the definitions and essential features of HCAI discussed in recent literature, e.g., \cite{Kaluarachchi-etal-2021-review,shneiderman2020,shneiderman2022human,wu2021survey,zhu2018value,robertson2020if,holstein2019improving}, we suggest the following working definition of HCR for:

\begin{definition*}
\emph{Research is human-centric if it satisfies the following condition: human stakeholders, in addition to the researcher, actively participate in the research project.}
\end{definition*}


To further clarify how stakeholders can be involved, we distinguish between four different pipeline stages within a research project:

\begin{itemize}

\item \textbf{Problem Definition.} The goal is to study the real needs of target users---rather than relying on the intuition of the researchers---and identify the gap between existing methods and user demands \cite{liao2020questioning,hong2021planning}. This then serves as the basis to formulate the research problem.

\item \textbf{Data Collection.} Either external stakeholders are directly involved in the data collection from planning to execution \cite{nekoto-etal-2020-participatory}, or the researchers use insights from the external stakeholders, which were obtained in the problem definition stage, to set up the data collection process. 

\item \textbf{Model Building.} There are two possible ways in which external stakeholders can impact the model building stage. (1) stakeholders may be involved before or after the training process; (2) stakeholders can be involved during model inference stage \cite{Kulesza2015, smith2018closing, Koh2017blackbox}.

\item \textbf{Task Evaluation.} 
Instead of---or in addition to---using automatic leaderboards or crowdsourced evaluations whose task is defined by researchers, systems are evaluated given requirements defined by external stakeholders \cite{ehsan2021expanding,lakkaraju2022rethinking}.
\end{itemize}

Given the definition for HCR and by distinguishing different research stages, we can discuss different opportunities and possibilities for HCR in a more structured manner.
\section{Guiding Questions in Research Stages}\label{sec:stages}
For each research stage we now formulate a series of guiding questions that can inspire future HCR.

\subsection{Stage 1: Problem Definition}\label{sec:pd}
Many research projects are not human-centric, because they do not take into account real human needs at inception; rather, they reflect the researcher's personal intuition and interest. Previously human-centric inspired NLP problems were for example machine translation \cite{weaver} or information extraction \cite{wilks1997information}, which were either defined or driven by real human need. Recent times have seen the emergence of more human-centric NLP tasks, such as the detection of hate speech on social media \cite{basile2019semeval} or fact checking \cite{mihaylova2019semeval}. These tasks are arguably human-centric---because they solve problems inspired by real human need---yet, they still typically lack the active involvement of the human stakeholders in the problem definition. To address this gap, we formulate four blocks of guiding questions that can help to make the problem definition of a research project more human-centric.

\paragraph{Guiding Questions.} 
\begin{itemize}
    \item Selection: Which NLP task will be tackled? What methods should be used?
    \item Stakeholders: Who are the stakeholders, what are their characteristics?  Who are the end users most affected by the system? Who operates the system, who owns it? Who shall be the target stakeholder?
    \item Pain points: What are the pain points for the target stakeholder?
    \item Research question: What are the shortcomings of existing methods? What research question should we ask to address these shortcomings in a research project?
\end{itemize}
Researchers often start by choosing an NLP task of interest and identifying gaps in research, which they would be interested in addressing. Typically the research is now conducted by the researchers without any additional outside input. In contrast, in a more human-centric view, researchers could first identify and involve stakeholders to derive a problem definition that would directly benefit the stakeholders by addressing their pain points in the area of research identified by the researchers. By this early involvement, the researchers can ensure that their research can have a direct positive impact.

Once a pain point to be addressed is defined, researchers should determine how well existing technology is able to address the defined problem. For this, the researchers, as the experts of the NLP domain, should analyse (1) which NLP task is most suitable and (2) what existing methods are a good starting point to solve the problem. Regarding the task selection, the researchers need to understand the problem as described by the other stakeholders and then to match it to the most suitable NLP task. This could for example be sequence classification, token classification, sequence generation or another task. Once the task is chosen, the researchers should explore promising existing methods which can solve the task. Here it is important that the researchers consider all viable options and that they do not restrict themselves to methods that they have previously used or know well.

\subsection{Stage 2: Data Collection}\label{sec:dc}
The design of training datasets as well as systems that support data collection, are crucial for creating effective human-centric NLP systems. For example, the standard crowdsourcing procedures could be intuitively considered as human-centric, because the data is collected with human input from non-researchers. Such crowdsourcing tasks, however, are defined by the researcher. With these constraints, neither the crowdworkers nor other external stakeholders (such as end users) can express their pain points outside of the limitations imposed by the researchers. Therefore, the process lacks creative human input from non-researchers, with which more meaningful data could be collected. 

Prior work on human-centric data collection includes (1) system that facilities domain knowledge acquisition from domain experts \cite{park2021facilitating}, (2) collection process that is gamified, e.g.~by asking users to generate text in a role playing game \citep{akoury-etal-2020-storium}, or (3) systems for machine-human co-creation \cite{clark2018creative,iskender-etal-2020-towards,tintarev2016personal}. The above examples demonstrate that NLP data generation tasks can benefit from creative and human-centric implementation of the data collection stage. The setup of future human-centric data collection projects can be facilitated by the following guiding questions.

\paragraph{Guiding Questions.}
\begin{itemize}
    \item Data: What data needs to be collected to solve the pain points of the end users?
    \item Annotation: Who is qualified to annotate? 
    \item Collection Approach: How should the annotation be done?
\end{itemize}

To address the above questions, ensure that all important angles are considered as well as offer transparency w.r.t.~to the collected data for all stakeholders, we recommend the guide on creating datasheets for new datasets by \citet{datasheets}.


\subsection{Stage 3: Model Building}\label{sec:mb}
Model building includes (1) defining the model and the training process, and (2) specifying the inference strategy. Typically this is done at the prerogative of the researcher, but both components can be made more human-centric. For example, model training could involve stakeholders during or after training. Overall, the type of human-machine interaction during the model building shapes whether it is a human-in-the-loop \cite{smith2018closing, yuksel2020human, khashabi2021genie} or machine-in-the-loop~\cite{clark2018creative}. The former means the user intervenes with the system operation, and the latter happens when the system’s support is triggered by the user to finish a  task.

During inference time, humans can provide inputs to assist the model with predictions (e.g. \citet{wen2019human,rello2015spellchecker}). In the context of explainable systems, stakeholders can for example be involved to 
ensure that a model follows certain explanations \cite{zhang2016, Zaidan2007, Arous2021}. Whether either type of involvement is desired, should be addressed in a discussion with all stakeholders.

\paragraph{Guiding Questions.}
\begin{itemize}
    \item Prototype: Can the designed model fulfil the external stakeholders' expectations?
    \item Inference: Should a stakeholder be involved in the inference process of the model?
\end{itemize}

To explore the answer to both questions, we recommend the researchers to repeatedly interact with the stakeholders in order to refine the prototype's architecture and to understand how end users later want to interact with the model. During the first iterations, one option could be to create simple mock-ups, e.g.~in form of slides that showcase the functionality and planned end user interaction. Once both questions can be answered sufficiently, the researchers can then start to build the system. Once a system is built and reaches an acceptable performance from the researchers' point of view, the researchers can start a continuous feedback loop between this stage and the next stage: task evaluation with the stakeholders.

\subsection{Stage 4: Task Evaluation}\label{sec:te} NLP systems are often evaluated on standard NLP benchmarks using scores averaged across a test set, such as accuracy (\textsc{GLUE} \cite{glue},  \textit{inter alia}). While such automatic benchmarking provides quick evaluation---and, consequently, speeds up NLP research---such a metric may fail to capture the diverse needs of the end users \cite{ethayarajh2020utility,narayan2021personalized}.

To address the issue of averaged metrics, \citet{ribeiro-etal-2020-beyond} propose to check systems against different types of examples. Other efforts in this direction include carrying out interviews to study NLP system failures \cite{hong2021planning} or investigating human debugging of NLP models \cite{lertvittayakumjorn2021explanation}. While these works move into the right direction, they do not explicitly capture human reactions from stakeholders, such as whether a system meets expectations or which open pain-points remain and what future wishes exist. For this, we formulate the following guiding questions.
\paragraph{Guiding Questions.}  
\begin{itemize}
    \item Metric: What is a good evaluation measure?
    \item Evaluator: Who evaluates the built model?
    \item Level: At which level is the task evaluated?
    \item Impact: What is done after the evaluation?
\end{itemize}

Especially when a human evaluator is involved, it is important to carefully investigate and define an appropriate evaluation metric (1st question). Together with external stakeholders, the researchers should determine the criteria for assessing if a stakeholder's pain point is addressed by the model. 
Once a metric is defined, all stakeholders should also address the question whether the evaluation task has been set up in a way that the evaluator can understand the task at hand. 

With regards to the 2nd and 3rd questions, the final goal should be that a target stakeholder evaluates the model on a task that aims to address their pain point(s). However, jumping straight to this setting may not be practical in most situations. Therefore, following \citet{doshivelez2017rigorous}, we propose an overall 3-step process for evaluation: (1) \textit{Non-HCR}: perform an automatic evaluation to ensure the model output is reasonable. Since human evaluators can be a rare and/or costly resource, an automatic evaluation provides a baseline level for quality. (2) \textit{Towards HCR}: recruit a human evaluator and evaluate on a pseudo-task. Due to time and cost constraints, it is often easier to find lay users (instead of e.g. a doctor) for evaluation. Additionally, if the real task is very complex, then a pseudo-task is helpful as a stepping stone; (3) \textit{HCR}: ask the targeted stakeholder to evaluate the model built to solve their pain point(s).

Finally, researchers can observe or interview end users to understand if the model addresses their pain points, fulfils their needs etc. This aids to address the 4th question: the insights can be used in an iterative way to improve the model building stage. Overall, the findings from the test stage can enrich the evaluation of research projects with a new dimension: stakeholders' feedback.

\section{Case Study: Text Summarization}\label{sec:case-study-summarization}

\begin{table*}[h!]
\centering
\begin{tabular}{|l|m{105mm}|} 
\hline
\multicolumn{1}{|c|}{Research Stages}  & \multicolumn{1}{c|}{Guideline Questions} \\ 
\hline
HCR.1. Problem Definition & 
1.1.~What is the users' pain points and can it be solved with summarization?
1.2.~Which end user needs to read which type of text? 
1.3.~Which users' needs are not covered by the existing techniques? 
1.4.~Why do the end users need a summary?
1.5.~What do they do with the summary?
1.6.~What do they need in a summary?\\ 
\hline
HCR.2. Data Collection & 
2.1.~Does suitable data already exist or does it need to be annotated? 
2.2.~Does the annotation (summary) have to be written by the end user or can the task be outsourced to other annotators?
2.3.~What is an effective way to collect all information end users need in a summary?\\ 
\hline
HCR.3. Model building  & 
3.1.~Should the end users correct the summarization of some sentences during the training process?
3.2.~Should the end users be able to select their summarization from a series of suggested snippets?\\
\hline
HCR.4. Evaluation & 
4.1~Can we improve end users decision speed or accuracy by giving them a summary?
4.2.~How can it be determined if a summary was helpful? 
4.3.~What needs to be in a summary to address end users' pain points?
4.4.~What does the stakeholder want to do based on the summary? Can they do this with the given summary?\\ 
\hline
\end{tabular}
\caption{Guideline questions adapted as example questions for the case study of summarization.}
\label{table:usecase_summarization}
\end{table*}

For each research stage, we provided a set of guiding questions to help guide future human-centric NLP research. In particular, by adapting the provided questions for specific tasks and addressing them, researchers can explicitly state in which sense their research project is human-centric. Based on this, we provide a case study on how the task of text summarization can be enriched to be more human centric.\footnote{While ideally concrete research questions are defined by interacting with external stakeholders, the case study and adapted questions are meant to serve as starting point for researchers to explore possibilities within HCR.} For additional case studies, see the appendix.

Text summarization by itself is not necessarily a HCR task. Whether or not a summarization project is human-centric depends on the defined scope of the project. For example, the commonly used dataset CNN / Daily Mail was created because the data was easily available: it consists of news text and corresponding summary bullets \cite{nallapati-etal-2016-abstractive}. Evaluating a model trained on this dataset with humans also does not make the project human-centric if the evaluation protocol is determined by the researcher and not the external stakeholders. In contrast, \citet{hsu-tan-2021-decision} take a step in the right direction: they define their research goal as finding a summary that supports human decision making. However, they also do not involve external stakeholders as active participants in the research project. To make this research project human-centric, one could for example identify an end user group, using interview techniques inspired by design thinking methods \cite{lewrick2020design}. A research question could then be: ``\textit{How might we help [a lawyer] to [quickly understand whether a given law text will help them with their current query]?}'' We list other possible questions in Table \ref{table:usecase_summarization}.

\section{Conclusion}
Human-centric research (HCR) can guide research output that benefits humans and addresses their pain points. We provided a working definition of what it means to do HCR and distinguished four different research stages where HCR components can be added. 
For each research stage, we discussed existing NLP research with HCR components and formulated a series of guiding questions, which can provide a starting point for future HCR research. The Appendix provides some initial ideas on how to explore HCR for 4 different NLP tasks.  In the future we want to further develop this framework to give more concrete ideas on how the guiding questions can be addressed in a research setting. 


\bibliography{hcrgraph}
\bibliographystyle{acl_natbib}

\clearpage
\appendix

\section{Case Studies}\label{app:case}
For each research stage, we provided a set of guiding questions to help guide future human-centric NLP research. In particular, by adapting the provided questions for specific tasks and addressing them, researchers can explicitly state in which sense their research project is human-centric. Based on this, we provide a series of case studies on how a particular NLP task can be enriched to be more human-centric.\footnote{While ideally concrete research questions are defined by interacting with external stakeholders, the case studies and adapted questions are meant to serve as starting point for researchers to explore possibilities within HCR.}

\paragraph{Case Study: Explainability}

\begin{table*}
\centering
\begin{tabular}{|l|m{105mm}|} 
\hline
\multicolumn{1}{|c|}{Stages}  & \multicolumn{1}{c|}{Guideline Questions}\\ 
\hline
HCR.1. Problem Definition  & 
1.1.~Who are the end users with which types of expertise, background and experience? 
1.2.~What are pain points of the end users such that explanations are needed?
1.3.~Which messages should the explanations include? 
1.4.~Which detail level of explanations is expected by the end users? 1.6.~What is the gap between existing techniques and the users' demands?\\
\hline
HCR.2. Data Collection & 
2.1.~Are there existing datasets publicly available or provided by the users/domain experts? 
2.2.~Does the annotation (explanation) have to be written by the end user or can the task be outsourced to other annotators?
2.3.~If outsourcing is allowed, how can we align annotators' explanations with the expectations of end users?
2.4.~What is an effective way to gather explanations that are meaningful and accurate?\\ 
\hline
HCR.3. Model building  & 
3.1.~Do the stakeholders need a self-explainable system or explanations for their legacy system?
3.2.~What is the time requirement of the explanations, immediate or deferred response?
3.3.~Should the model provide explanations in an interactive mode?
3.4.~Would it be beneficial if the end user is involved in generating the explanation and/or the prediction?\\ 
\hline
HCR.4. Evaluation & 
4.1.~How can it be determined if an explanation was helpful? 
4.2.~What needs to be in an explanation to address end users' pain points?
4.3.~Do we need to measure meaningfulness, accuracy and completeness of explanations and how can this be done?\\ 
\hline
\end{tabular}
\caption{Guideline questions adapted as example questions for the case study of explainability.}
\label{table:usecase_explainability}
\end{table*}

Explainable NLP has attracted increasing attention, as most state-of-the-art NLP models (e.g. deep neural networks) are perceived as complex ``black-boxes''.
By definition, explainability research should intrinsically be human-centric, since the goal is to provide explanations that are understandable and meaningful for users of an AI system. However, in practice, most of existing works investigate explanations driven by the 
algorithmic view of the researchers \cite{ehsan2021expanding}, rather than the real needs of the end users, and thus these works would not necessarily be human-centric. 
For example, a line of typical explainable NLP research is to highlight features (tokens and sentences) of input texts to explain the predictive results. Some algorithmic metrics derived with complicated math, e.g. area over the MoRF perturbation curve (AOPC), are used to measure the performance \cite{samek2016evaluating}. Sophisticated models are developed to pursuit higher measurement scores \cite{chen2020variational,lei2016rationalizing}. Although there exist some works employing human to evaluate the explanations, e.g., \citet{pruthi-etal-2020-weakly}, the annotation protocol is designed to match the researchers' understanding about the explanations, rather than the demands of the end users the NLP system will serve. Some recent works have realized the gap. \citet{shen2020interpretation} demonstrate that saliency maps, a commonly used explanation method, actually decreases the capability of the general users to figure out why a deep neural networks makes an error. \citet{phillips2020principlesXAI} propose four principles of XAI that emphasize meaningfulness of explanations for different groups of end users, and point out there is no ``one-fits-all'' explanation. To improve user experiences with XAI, \citet{liao2020questioning} propose XAI Question Bank by interviewing 20 AI practitioners. 
The major gap between existing explanations and human-centric ones is due to lack of external stakeholders as active participants in the research projects.

Based on the guideline questions introduced in the paper, we formulate a set of questions specifically for explainability research in Table~\ref{table:usecase_explainability}, which we hope can help researchers to position their work as HCR and helps to communicate how their research is human-centric. 

\paragraph{Case Study: Information Extraction}
Information extraction (IE) aims to draw out structured information, such as entities and relations, from unstructured textual data. Compared with other NLP tasks, e.g.~natural language parsing and POS tagging, IE was previously viewed as a human-centric inspired NLP problem, as it was driven by real human need \cite{wilks1997information}, yet it still typically lacks the active involvement of the external stakeholders. Most research focuses on the researcher's intuition and interest, e.g., Open IE \cite{kolluru2020IMoJIE,cui2018openIE,gashteovski2017MinIE,gashteovski2019opiec} and scalable IE \cite{lin2020KBPearl}.

Recent literature starts including other humans in the IE projects. For example, \citet{zhang2019EntityExtraction} involve humans in entity extraction for data annotation. In particular, human annotators are employed to formulate regular expressions that generate weak labels to train an initial model, then they manually label the texts to refine the pretrained model. 
\citet{gashteovski-etal-2020-aligning} integrate human knowledge to evaluate alignment between the Open IE triples and the DBpedia KB w.r.t. information content. 
Again, most of the existing IE projects that employ humans in the research stages neither actively analyze the distinct needs of different groups of end users, nor clearly integrate external stakeholders' view and intelligence into different research stages. So they are not necessarily human-centric research.

Based on the guideline questions introduced in the paper, we formulate a set of questions specifically for information extraction research in Table~\ref{table:usecase_IE}, which we hope can help researchers to position their work as HCR and helps to communicate how their research is human-centric. 
\begin{table*}
\centering
\begin{tabular}{|l|m{105mm}|} 
\hline
\multicolumn{1}{|c|}{Stages}  & \multicolumn{1}{c|}{Guideline Questions} \\ 
\hline
HCR.1. Problem Definition  & 
1.1.~Who are the end users? 
1.2.~Which types of entities, relations, conditions and events should be extracted from which category of texts?
1.3.~Once extraction is done, what is done with the extracted information?
1.4.~What is the typical daily routine of the end users with IE?
1.5.~What is the IE related pain point of the end users?
1.6.~What is the gap between existing techniques and users' demands?\\
\hline
HCR.2. Data Collection & 
2.1.~Are there existing datasets or related annotations? 
2.2.~Does the annotation have to be done by the end users or can be outsourced to other annotators?
2.3.~Are there existing annotation tools matching the requirements of the end users?\\ 
\hline
HCR.3. Model building  & 
3.1.~Can the model integrate the experience/knowledge of the end users to reduce complexity but keep or even increase flexibility?
3.2.~Does the model extract information in a batch mode, or an interactive mode?
3.3.~Would it be beneficial if the end users are involved in training and inference?\\ 
\hline
HCR.4. Evaluation & 
4.1.~How can it be determined if the extracted information was helpful? 
4.2.~How can we measure the effectiveness of the extracted information to address end users' pain points?\\ 
\hline
\end{tabular}
\caption{Guideline questions adapted as example questions for the case study of information extraction.}
\label{table:usecase_IE}
\end{table*}

\paragraph{Case Study: Document Similarity}
Semantic Textual Similarity (STS) is an important NLP task, which is widely applied in various fields such as business industry, geoinformation and biomedical informatics \cite{majumder2016}. The increasing popularity of STS in realistic scenarios motivates the research and development of STS technology to be human-centric. 

However, with the availability of the benchmark dataset such as STS-B \cite{cer2017} and the open-source leaderboards
, STS research is often motivated by the purpose of beating the state-of-art algorithm rather than the consideration of realistic needs of the human users. 


Based on the guideline questions introduced in the paper, we formulate a set of questions specifically for document similarity research in Table~\ref{table:usecase_document_similarity}, which we hope can help researchers to position their work as HCR and helps to communicate how their research is human-centric. 


\section{Limitations}
Human-centric research in NLP is still a young research field. We aim to provide a first working definition for the term, we state how NLP research can be split into different stages and proceed to address how future research projects could incorporate more human-centric components into the different stages. 
We acknowledge that there are also other ways in which research projects can be made human-centric compared to the suggestions provided in this paper. 
We would also like to point out that our statements and guiding questions in this paper are not cut in stone but rather should provide a solid base for discussions and comparisons to other approaches. As this is still a young research field, we think it is important to encourage such discussions at an early stage.
With this paper, we hope to provide a good basis for future discussions with the community as well as inspiration for other researchers who might want to conduct more human-centric research.

\begin{table*}
\centering
\begin{tabular}{|l|m{105mm}|} 
\hline
\multicolumn{1}{|c|}{Stages}  & \multicolumn{1}{c|}{Guideline Questions}\\ 
\hline
HCR.1. Problem Definition  & 
1.1.~Who are the end users with which types of expertise, background and experience (e.g. caseworkers, online customers)? 
1.2.~What are the pain points of the currently in-used similar document searching method (e.g. too slow, not precise enough)?
1.3.~Can we use the semantic textual similarity system to solve the pain point (e.g, can we use our STS system to quickly and precisely search the similar cases from the archive database)?\\
\hline
HCR.2. Data Collection & 
2.1.~What kind of data (e.g. text, tabular) and labels (e.g. binary label, float similarity score) is needed to build a specific STS model? 
2.2.~Is the data available (e.g. public dataset, private dataset)?
2.3.~Is the additional annotation needed?
2.4.~Who is qualified to define the similarity scoring (e.g. researchers, domain experts, standard regulations)?
2.5.~Who is allowed to annotate the labels (e.g. crowdsource annotators, domain experts, intelligent annotating system)?
2.6.~Is a human-machine collaborative method feasible by considering the human capacity and technical abilities?\\ 
\hline
HCR.3. Model building  & 
3.1.~What extra features do stakeholders expect (e.g. the similarity estimation model could be expected to be able to handle multi-modal data)? 
3.2.~What is the requirement of end users on the model execution (e.g. learning time, computational cost and progress monitor)? 
3.3.~What kind of output is needed (e.g. a list of top-k similar documents, rank all documents via their similarity scores)?
3.4.~Does the training process involve human interference (e.g. adjust learning parameter)? 
3.5.~Would it be beneficial to involve end users in inference? Who is qualified to be involved?\\ 
\hline
HCR.4. Evaluation & 
4.1.~How would the end users define if two documents are similar?
4.2.~How would the end users evaluate the accuracy of the searching results?
4.3.~Which stakeholder is qualified to define the evaluation metrics?
4.4.~What do the end users plan to do with the result of the system? How can we measure whether they succeeded based on the STS model?\\ 
\hline
\end{tabular}
\caption{Guideline questions adapted as example questions for the case study of document similarity.}
\label{table:usecase_document_similarity}
\end{table*}

\end{document}